\RequirePackage{fix-cm}
\documentclass[letterpaper,10pt,conference]{ieeeconf}
\IEEEoverridecommandlockouts
\usepackage[T1]{fontenc}
\usepackage[utf8]{inputenc}
\usepackage{graphicx}
\usepackage{wrapfig}
\usepackage{amsmath,amssymb}
\usepackage{booktabs,array,tabularx}
\usepackage{multirow}
\usepackage{threeparttable}
\usepackage{xcolor}
\usepackage{xurl}
\usepackage{listings}
\IfFileExists{inconsolata.sty}{\usepackage{inconsolata}}{\usepackage{courier}}
\usepackage{microtype}
\usepackage{balance}
\usepackage[hidelinks]{hyperref}
\hypersetup{pdftitle={MobileCopilot: A Domain-Specific Language Framework for Agentic Mobile Robot Design},pdfsubject={MobileDSL and compiler-guided mobile robot development},pdfauthor={Akihisa Watanabe, Jiawei Fang, Zhiyang Dou, Yuxin Ray Song, Zizhao Tong, Yunsheng Tian, Minghao Guo, Benjamin Jones, Wojciech Matusik}}
\newcommand{\AgentModel}{gpt-5.6-sol}
\newcommand{\AgentName}{Codex}
\newcommand{\AgentBudget}{1800}
\newcommand{\NumCases}{50}
\newcommand{\AblationAgent}{Codex}
\newcommand{\AblationModel}{gpt-5.6-sol}
\makeatletter
\newcommand{\CompactTableCaption}{%
  \long\def\@makecaption##1##2{%
    \vspace*{2pt}%
    {\normalfont\fontsize{8}{9.5}\selectfont\raggedright
    \noindent ##1.\enspace ##2\par}%
    \vskip 3pt}%
}
\makeatother

\definecolor{codebackground}{RGB}{246,248,250}
\definecolor{codekeyword}{RGB}{157,34,50}
\definecolor{codestring}{RGB}{28,59,96}
\lstdefinestyle{mystyle}{
  backgroundcolor=\color{codebackground},
  keywordstyle=\color{codekeyword},
  stringstyle=\color{codestring},
  commentstyle=\color{gray},
  basicstyle=\fontsize{7.3}{8.4}\selectfont\ttfamily,
  breaklines=true,breakatwhitespace=false,
  columns=fullflexible,keepspaces=true,
  showstringspaces=false,tabsize=2,
  frame=tb,numbers=left,numberstyle=\tiny\color{gray},
  numbersep=4pt,xleftmargin=0pt,
  linewidth=\columnwidth,
  aboveskip=6pt,belowskip=3pt,captionpos=b
}
\lstdefinelanguage{robotdsl}{morekeywords={Robot,Mecanum_four_wheel,Mecanum,Camera,Lidar,Arm},sensitive=true,morestring=[b]",morecomment=[l]{//}}
\lstdefinelanguage{worlddsl}{morekeywords={World,Scenario,Model,Goal,spawn},sensitive=true,morestring=[b]",morecomment=[l]{//}}
\lstdefinelanguage{taskdsl}{morekeywords={Task,PipelineSequence,ComputePathToPose,FollowPath},sensitive=true,morestring=[b]",morecomment=[l]{//}}
\lstdefinelanguage{evaluatedsl}{morekeywords={Evaluate,Check_vel,Check_acc,Check_distance,Check_clearance,Check_time},sensitive=true,morestring=[b]",morecomment=[l]{//}}
\lstdefinestyle{robotdsl}{style=mystyle,language=robotdsl}
\lstdefinestyle{worlddsl}{style=mystyle,language=worlddsl}
\lstdefinestyle{taskdsl}{style=mystyle,language=taskdsl}
\lstdefinestyle{evaluatedsl}{style=mystyle,language=evaluatedsl}

\makeatletter
\@ifpackageloaded{ulem}{\normalem}{\RequirePackage[normalem]{ulem}}
\makeatother
\RequirePackage{needspace}

\providecommand{\MCBeforeQuestion}[1]{}
\newcommand{\MCQuestion}[2]{%
  \MCBeforeQuestion{#1}%
  \par\addvspace{0.2\baselineskip}%
  \Needspace{5\baselineskip}%
  \noindent\textbf{Q#1: #2}\par\nobreak\vspace{1pt}}
\newcommand{\MCAnswer}[1]{%
  \noindent\begingroup
  \renewcommand{\ULthickness}{0.4pt}%
  \setlength{\ULdepth}{1.5pt}%
  \uline{A: #1}\par
  \endgroup\vspace{1pt}}

\providecommand{\MCResultsStart}{\par\Needspace{9\baselineskip}}

\title{\LARGE\bf MobileCopilot: A Domain-Specific Language Framework\\for Agentic Mobile Robot Design\vspace{-16pt}}
\author{Akihisa Watanabe$^{*,\dagger}$, Jiawei Fang$^{*,\dagger}$, Zhiyang Dou$^{*}$,
Yuxin Ray Song, Zizhao Tong,\\
Yunsheng Tian, Minghao Guo, Benjamin Jones, and Wojciech Matusik%
\thanks{$^{*}$These authors contributed equally.}%
\thanks{$^{\dagger}$Work done during an internship at MIT.}%
\thanks{Massachusetts Institute of Technology (MIT).}%
}
\newcommand{\paperteaser}{\begin{minipage}{\textwidth}
\centering
\includegraphics[width=\linewidth]{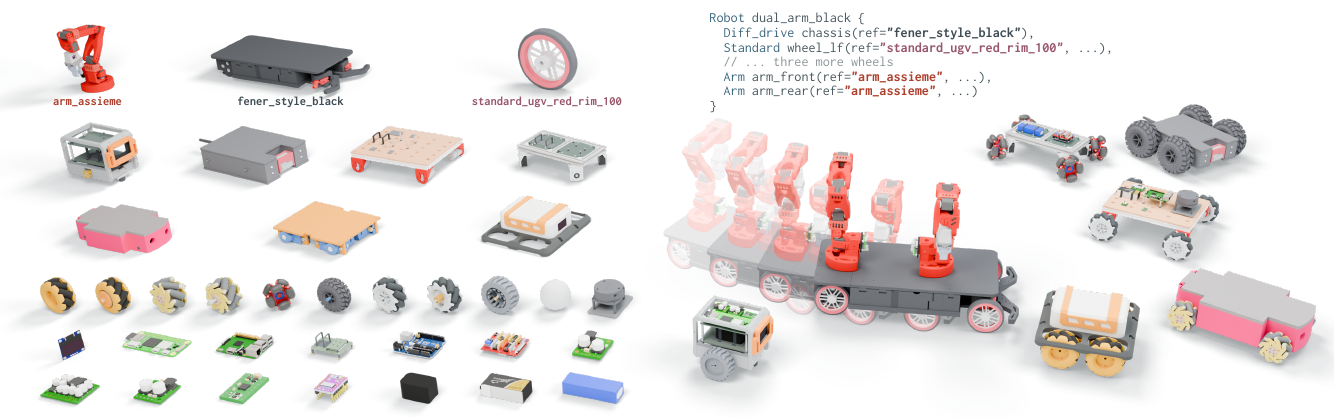}\\[-2pt]
{\footnotesize\raggedright
\textbf{MobileDSL: from reusable components to diverse mobile robots.}
Components from a shared library (left) are combined into different robot
designs (right). The MobileDSL code excerpt defines the dual-arm platform
using a chassis, four wheels, and two arms.\par}
\vspace{4pt}
\end{minipage}}
\IEEEaftertitletext{\vspace{-24pt}\paperteaser}
\begin{document}
\maketitle
\thispagestyle{empty}
\pagestyle{empty}
\begin{abstract}
Mobile robot development relies on separate descriptions of robot structure, environment, behavior, and evaluation, leaving developers to maintain consistency across them. We present \textit{MobileDSL}, a unified domain-specific language for mobile robot development that makes assembly and system integration part of compilation. The compiler assembles the robot and generates its controller and navigation settings from the same program, using shared references to check dependencies across the system. Design changes are reflected in the generated robot and its settings rather than repaired file by file. 
For efficient automated development, we introduce \textit{MobileCopilot}, an agentic framework that uses MobileDSL and its compiler as tools. Given a user request, a coding agent retrieves components, writes a program, and refines the design using compiler feedback. On a benchmark of 50 mobile robot design tasks, this workflow increases the proportion of systems that complete their programmed behavior in simulation by 42--50 percentage points over direct authoring of ROS/Gazebo files, while reducing API cost by approximately 33--61\% and requiring less code and generation time.

\end{abstract}
\section{Introduction}
\label{sec:intro}
Mobile robots are used in logistics~\cite{ramdani2019safe,bloss2011mobile}, healthcare~\cite{kebede2024review}, service applications~\cite{garcia2020robotics}, and exploration~\cite{garaffa2021reinforcement}. Developing these systems requires coordinating their physical structure, sensing, control, environment, behavior, and evaluation~\cite{raj2022comprehensive}. Yet these aspects are usually described separately: URDF~\cite{urdf} or MJCF~\cite{todorov2012mujoco} for robot structure, ROS~\cite{ros2009} for software integration, Gazebo~\cite{gazebo2025} for simulation, behavior trees~\cite{ogren2022behavior} for tasks, and application-specific scripts for evaluation~\cite{guiochet2017safety}.

The difficulty is not only the number of formats, but the need to express the same design decision in several places. Wheel dimensions and placements determine quantities used by both the robot model and its controller~\cite{tola2024understanding}. A navigation goal is used by both a behavior and its evaluation. Changing a wheel or moving a goal therefore requires corresponding updates wherever that information is used. Keeping these values synchronized adds work to each design iteration and creates opportunities for inconsistent settings.

Large language models (LLMs) can generate robot programs~\cite{zeng2022codeaspolicies,hu2024codebotler}. For coding agents, the available tool interfaces also shape the information and feedback used during revision~\cite{aizawa2025tools}. When robot structure, controller settings, and task references are authored separately, the agent must maintain dependencies across these descriptions as the design changes. Syntactic validity of the individual files does not establish consistency of the resulting system. This motivates a shared design representation that makes these dependencies explicit and supports compiler diagnostics in terms of components, world objects, and actions.

We introduce \textbf{MobileDSL}, a unified domain-specific language for mobile robot development. Its four language modules cover robot structure (\textit{CoreDSL}), environments (\textit{WorldDSL}), behavior (\textit{TaskDSL}), and evaluation requirements (\textit{EvaluateDSL}). They share references to robot components, world objects, and tasks. A navigation action and a distance requirement, for example, can name the same world goal, so changing that goal updates both uses through recompilation.

MobileDSL makes assembly and system integration part of compilation. The developer specifies components, connections, and behavior in one program; the compiler assembles the robot and generates its control and navigation configuration from the same design. A design change is reflected across the generated system rather than repaired file by file. The compiler also checks dependencies across robot structure, environment, and behavior, identifying incompatible components, missing world goals, and actions unsupported by the selected robot.

For efficient automated mobile robot development, we introduce \textbf{MobileCopilot}, an agentic framework that uses MobileDSL and its compiler as tools. Given a user request, the agent retrieves components, writes a MobileDSL program, and iteratively refines the design using compiler feedback.

We develop a benchmark of 50 mobile robot design tasks and an evaluation harness that runs the generated systems in simulation. Compared with direct authoring of ROS/Gazebo files, MobileDSL increases the proportion of systems that complete their programmed behavior by 42--50 percentage points, while reducing API cost by approximately 33--61\%.

Our contributions are:
\begin{enumerate}
\item MobileDSL, a unified language for robot structure, environment, behavior, and evaluation requirements, with a compiler that assembles components and keeps their generated settings consistent.
\item MobileCopilot, an agentic robot development framework that combines component retrieval and MobileDSL program generation with iterative refinement guided by compiler diagnostics.

\item A benchmark of 50 mobile robot design tasks and a shared evaluation harness for comparing system execution and agentic authoring effort.
\end{enumerate}

\section{Related Work}
\subsection{Domain-Specific Languages in Robotics}
Domain-specific languages support different aspects of robot development~\cite{mernik2005and,nordmann2014survey}. Common workflows use \mbox{URDF/Xacro} to describe robot structure~\cite{urdf,xacro,tola2024understanding}, ROS for integration~\cite{ros2009}, Gazebo or USD-based tools for simulation~\cite{koenig2004gazebo,nvidia2023openusd}, and behavior trees for task execution~\cite{colledanchise2018behavior,iovino2022survey}. RobotML jointly models robot structure and software~\cite{dhouib2012robotml}, while USD-based workflows and Isaac Sim organize scene representations and dynamics~\cite{nvidia2023openusd,isaac_sim2025}. MobileDSL focuses on the relationships between a mobile robot's morphology, environment, behavior, and evaluation. Its compiler aligns parts at their mounting points and derives control settings from the resulting geometry, keeping related descriptions consistent.

\subsection{LLM-Assisted Robot Development}
LLM-assisted robotics includes both programming existing robots and designing new ones. Code-as-Policies~\cite{zeng2022codeaspolicies} and CodeBotler~\cite{hu2024codebotler} generate control programs for existing robots. RoboGrammar uses graph grammars for terrain-aware morphology design~\cite{zhao2020robogrammar}; Text2Robot~\cite{ringel2025text2robot}, Evolution 6.0~\cite{khan2025evolution}, and LASeR~\cite{song2025laser} explore language-guided robot design. RoboMoRe couples morphology and control-related objectives~\cite{fang2025robomore}. Our focus is on assembling reusable mobile-robot components and connecting the resulting robot to its environment, task, and evaluation. MobileCopilot gives the language model access to assembly, configuration, and validation tools, so it can revise the design using constraints that span the complete system.

\section{MobileDSL}
\label{sec:mobiledsl}
MobileDSL represents a mobile robot and its operation as one program composed of four language modules. Shared names connect components, world objects, actions, and evaluation requirements. The examples below describe a mecanum robot navigating to a goal, together with bounds on speed, distance, clearance, and completion time.

The compiler resolves references, checks compatibility, and derives assembly transforms and settings. A backend translates the result into files for the target execution environment, separating language analysis from target-specific code generation.

\subsection{CoreDSL}
\begin{figure}[t]
\centering
\setlength{\abovecaptionskip}{2pt}
\includegraphics[width=\linewidth,trim=0 18bp 0 0,clip]{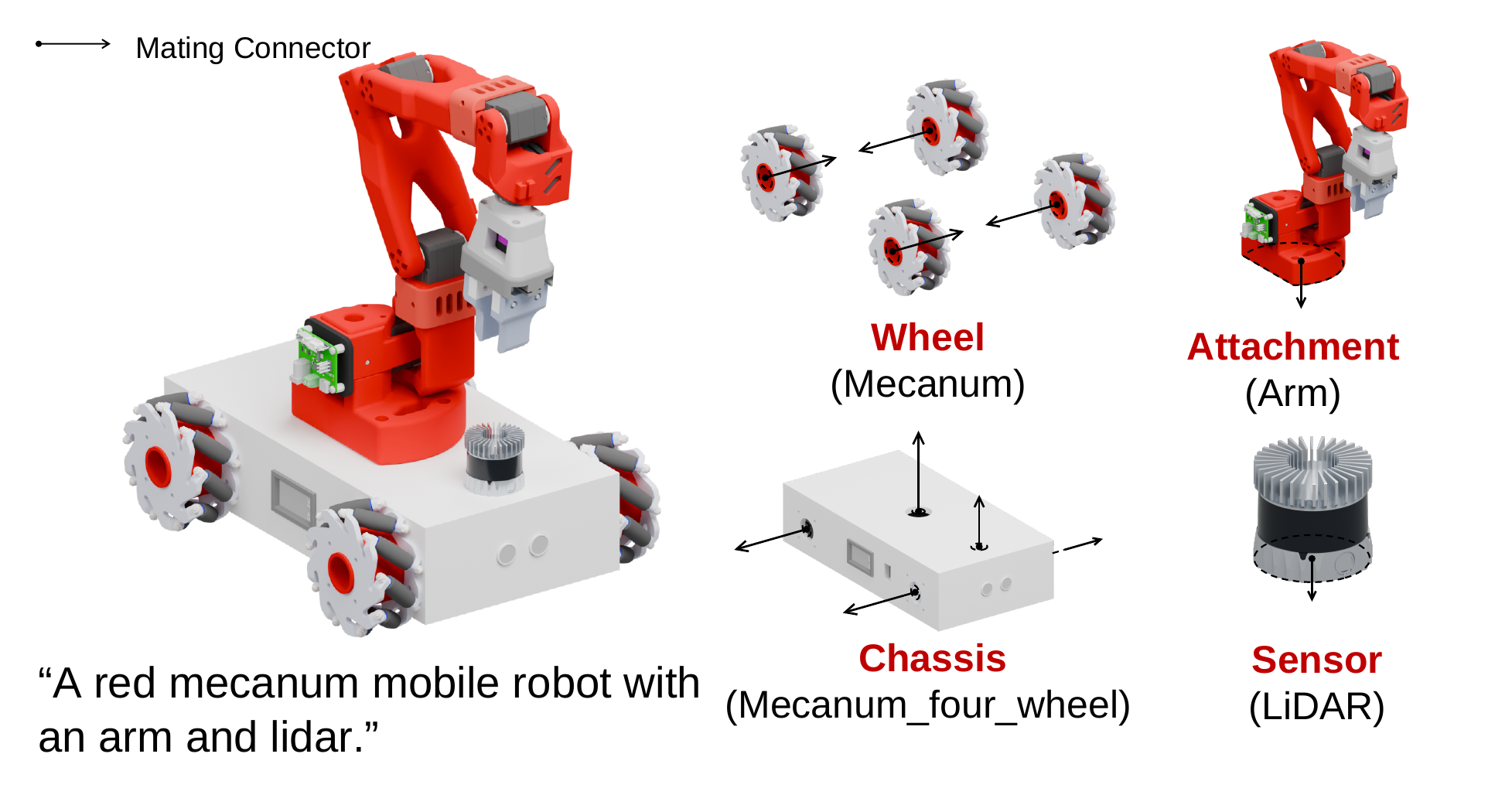}
\caption{\textbf{CoreDSL component roles and assembly.} A mecanum base, front-mounted lidar, and centrally mounted arm illustrate how chassis, wheels, sensors, and attachments connect through mating connectors.}
\label{fig:example}
\end{figure}

CoreDSL expresses a robot as an assembly of reusable components. It combines \emph{functional component roles}, \emph{assembly through mounting points}, and \emph{automatic configuration}. Figure~\ref{fig:example} illustrates the component roles and their connections. DSL~\ref{lst:coredsl} begins a navigation example with the illustrated four-wheel mecanum robot, a front-mounted lidar, and a centrally mounted arm.
\par\vspace{4pt}\noindent\begin{minipage}{\linewidth}
\begin{lstlisting}[style=robotdsl,caption={CoreDSL: a mecanum robot with lidar and an arm.},label={lst:coredsl}]
Robot nav_bot {
  Mecanum_four_wheel chassis(ref="silver"),
  Mecanum wheel_lf(ref="red",glue="0.0976 0.095 -0.0443 0 1 0"),
  Mecanum wheel_lb(ref="red",glue="-0.1124 0.095 -0.0443 0 1 0"),
  Mecanum wheel_rf(ref="red",glue="0.0976 -0.095 -0.0443 0 -1 0"),
  Mecanum wheel_rb(ref="red",glue="-0.1124 -0.095 -0.0443 0 -1 0"),
  Lidar front_scan(ref="lidar",glue="0.11 -0.0001 -0.0043 0 0 -1"),
  Arm top_arm(ref="arm",glue="-0.0073 -0.0001 -0.0043 0 0 -1")
}
\end{lstlisting}
\end{minipage}\par

\paragraph{Component Roles}
A generic link does not indicate whether it serves as a chassis, wheel, or sensor. CoreDSL assigns components these functional roles, distinguishing \textbf{chassis}, \textbf{wheels}, \textbf{sensors}, and \textbf{attachments} (Fig.~\ref{fig:example}). This information constrains how the robot is assembled and controlled. A four-wheel mecanum chassis, for example, requires a corresponding wheel arrangement and controller family, while allowing the individual wheel positions to vary.

\begin{figure}[t]
\centering
\includegraphics[width=\linewidth,trim=0 28bp 0 0,clip]{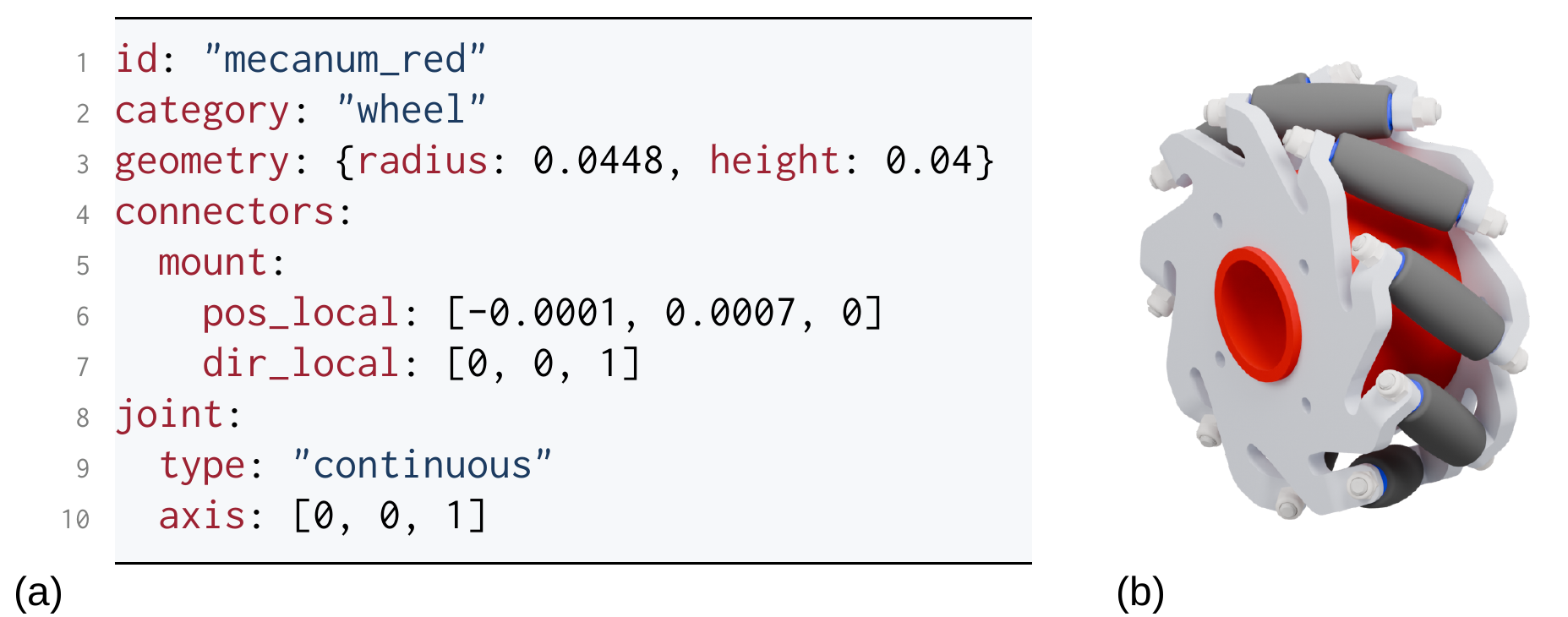}\\[-3pt]
\caption{\textbf{Reusable component description.} Descriptions and images support component selection; geometric parameters and connector frames specify how the components fit together.}
\label{fig:component-description}
\end{figure}

\paragraph{Connector-Based Assembly}
Each library component is described by its functional role, geometric parameters, joints, and connector frames (Fig.~\ref{fig:component-description}). The \texttt{ref} field selects a library component. The \texttt{glue} tuple specifies the mating point and normal in the parent frame, with an optional roll angle. The compiler aligns the selected component's connector with this placement and checks its compatibility with the assembly. The user therefore specifies where components connect without separately calculating every link-to-link transform.

\begin{figure}[t]
\centering
\setlength{\abovecaptionskip}{2pt}
\includegraphics[width=\linewidth,trim=0 28bp 0 0,clip]{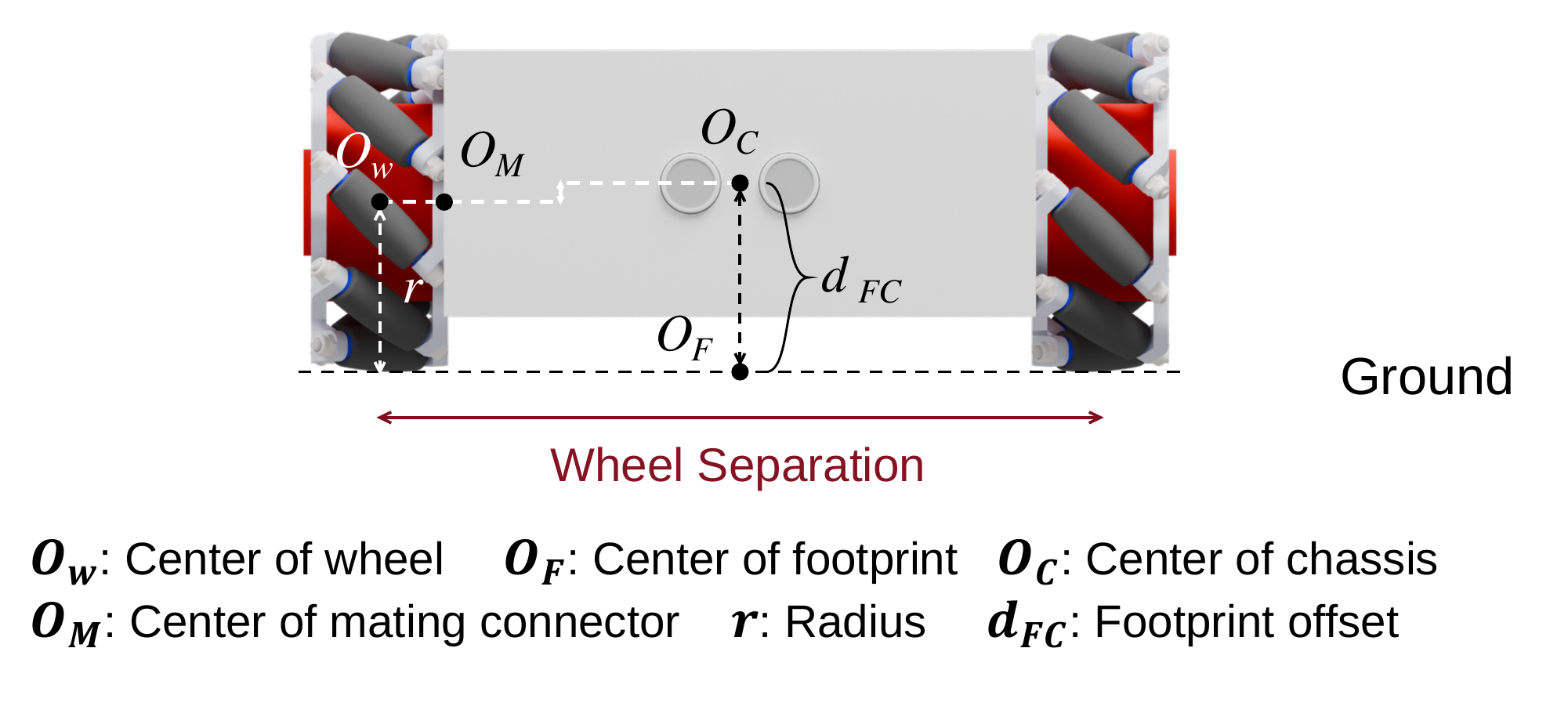}
\caption{\textbf{Configuration from geometry.} Wheel radius and aligned connector frames determine the footprint offset and controller geometry, including wheel separation and ordering.}\vspace{-6pt}
\label{fig:autoconfig}
\end{figure}

\Needspace{7\baselineskip}
\paragraph{Automatic Configuration}
The compiler uses the assembled geometry to derive controller and ground-reference settings. Let ${}^{C}T_M$ and ${}^{w}T_M$ denote the mating-connector poses in the chassis and wheel frames, respectively. Aligning the connectors gives the wheel pose in the chassis frame:
\begin{equation}
{}^{C}T_w = {}^{C}T_M\bigl({}^{w}T_M\bigr)^{-1}.
\label{eq:alignment}
\end{equation}
The wheel-center position ${}^{C}p_w$ and orientation ${}^{C}R_w$ follow from this transform. For the geometry in Fig.~\ref{fig:autoconfig}, a wheel of radius $r$ gives the chassis-to-ground footprint offset
\begin{equation}
d_{FC}=r-({}^{C}p_w)_z.
\label{eq:footprint}
\end{equation}
This offset locates the ground-referenced footprint frame. The same wheel positions, orientations, and radii determine controller parameters such as wheel ordering, separation, and wheelbase. The assembled geometry also determines the navigation footprint. Changing a wheel or its placement therefore updates these derived settings together, instead of requiring separate edits to the robot and controller descriptions.

\subsection{WorldDSL}
WorldDSL describes the environment, the robot's initial pose, and named objects or locations relevant to its task. It combines reusable scenarios with positioned models, including assets from Gazebo Models and Fuel~\cite{gazebo_models,gazebo_fuel}. DSL~\ref{lst:worlddsl} places the robot in a room with a navigation goal.
\par\vspace{4pt}\noindent\begin{minipage}{\linewidth}
\begin{lstlisting}[style=worlddsl,basicstyle=\fontsize{6.6}{7.4}\selectfont\ttfamily,numberstyle=\fontsize{5.5}{6}\selectfont\color{gray},aboveskip=2pt,belowskip=2pt,caption={WorldDSL: room, robot pose, and goal.},label={lst:worlddsl}]
World navigate_world {
  spawn { "@core/nav_bot": "0 0 0 0 0 0 1" }
  Scenario room(ref="living_room") { }
  Goal goal(pose="0 2 0 0 0 0 1")
}
\end{lstlisting}
\end{minipage}\par
TaskDSL actions and EvaluateDSL requirements refer to these entities by name. Changing a goal's position updates both the generated task coordinates and its distance-check target, avoiding separate copies of the same position.

\subsection{TaskDSL}
TaskDSL expresses behavior through a tree of navigation and locomotion actions using Nav2 primitives~\cite{macenski2020marathon2}. DSL~\ref{lst:taskdsl} computes a path to the WorldDSL goal and follows it; \texttt{\{path\}} passes the path between the two actions.
\par\vspace{4pt}\noindent\begin{minipage}{\linewidth}
\begin{lstlisting}[style=taskdsl,caption={TaskDSL: path planning and following.},label={lst:taskdsl}]
Task navigate_safely {
  PipelineSequence {
    ComputePathToPose(goal="@world/goal",path="{path}"),
    FollowPath(path="{path}")
  }
}
\end{lstlisting}
\end{minipage}\par
The compiler checks tree structure, action parameters, and referenced goals. It also checks whether actions match the selected robot: a spin is rejected when its drive configuration forbids in-place rotation. A \texttt{Parallel} node is rejected when multiple child branches can command locomotion. A navigation setup requiring laser scans must include a scan-producing sensor.

The compiler can also reject an insufficient task time limit. It derives a duration lower bound from explicit waits and supported constant-speed motions: sequential actions add their bounds, alternatives take the minimum, and overlapping pipeline actions take the maximum. Actions without a supported duration estimate, such as navigation, contribute zero. The compiler rejects a bound greater than the task's \texttt{timeout\_s}. This is a conservative consistency check on the declared task budget.

\begin{figure*}[t]
\centering
\setlength{\abovecaptionskip}{2pt}
\includegraphics[width=\linewidth]{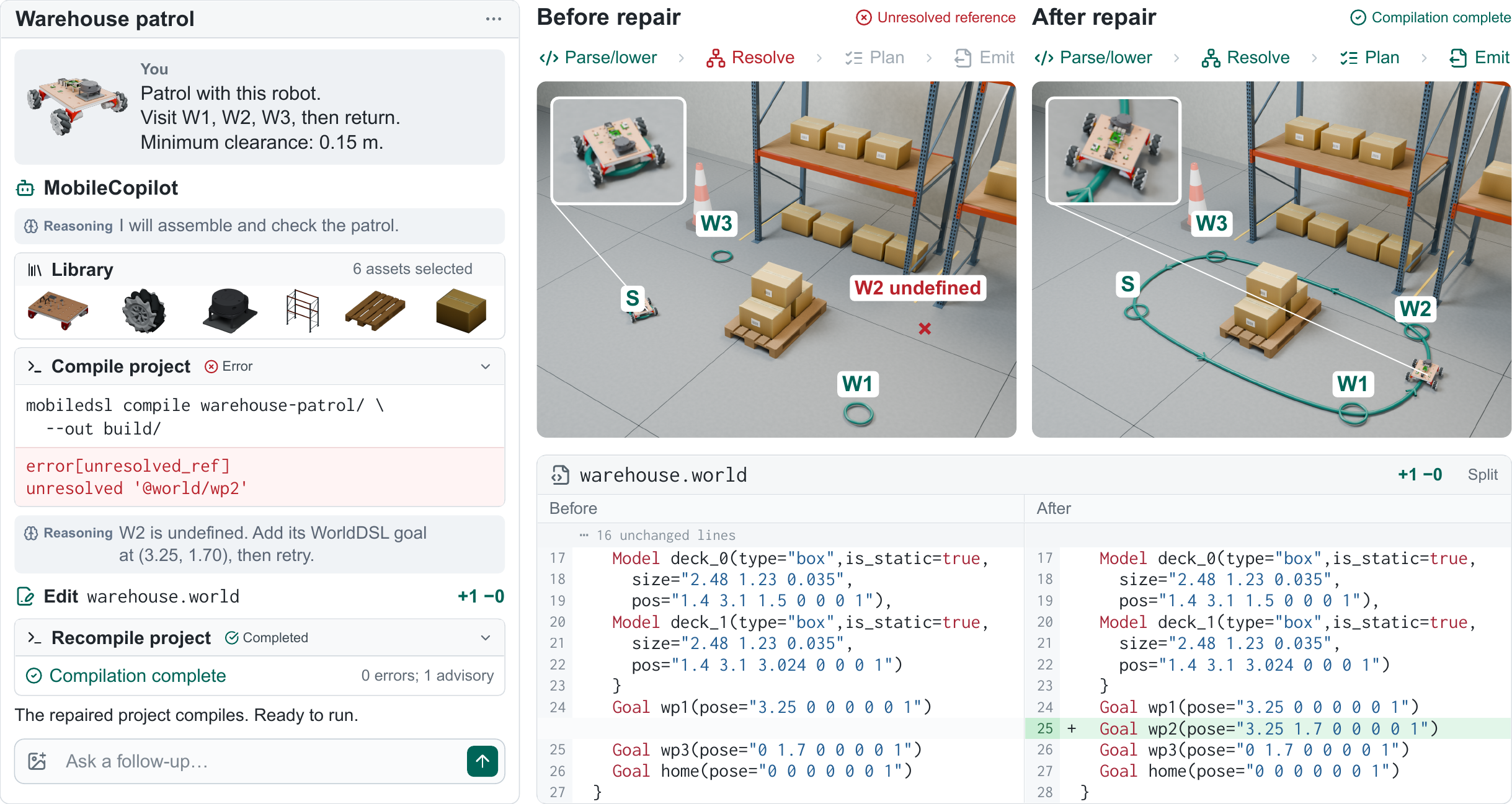}
\caption{\textbf{MobileCopilot workflow with compiler feedback.} The agent retrieves components, writes a program, and revises it using compiler diagnostics (left). Adding the missing WorldDSL waypoint (lower right) resolves the task reference and allows compilation to complete (upper right).}
\label{fig:pipeline}
\end{figure*}

\subsection{EvaluateDSL}
\label{sec:evaluate}
EvaluateDSL states what should be observed and which bounds should hold, using the same components, world objects, and tasks as the rest of the program. A developer can attach a speed limit to a wheel joint or a distance requirement to a named goal without repeating its coordinates or manually identifying the corresponding observation source.
\par\vspace{4pt}\noindent\begin{minipage}{\linewidth}
\begin{lstlisting}[style=evaluatedsl,caption={EvaluateDSL: speed, goal distance, clearance, and duration.},label={lst:evaluatedsl}]
Evaluate navigation_safety_check {
  Check_vel(target="@core/wheel_lf_joint",min=-12,max=12),
  Check_distance(target="@world/goal",max=0.15),
  Check_clearance(min=0.05),
  Check_time(target="@task/navigate_safely",max=30)
}
\end{lstlisting}
\end{minipage}\par
The compiler validates the target types and numerical bounds, then emits an evaluation plan containing the resolved targets, quantities, limits, and observation sources. Joint velocity and effort map to joint-state data, goal distance to simulator poses, clearance to the nearest lidar range, and duration to task timing. Unsupported observation bindings produce warnings. Lidar clearance measures proximity, distinct from full-body collision checking.

The resulting plan keeps observation targets and bounds linked to the robot program as components or world goals change. In this benchmark, EvaluateDSL plans are generated and checked statically but are not executed. All configurations are scored by the shared runtime evaluator described in Sec.~\ref{sec:benchmark}.

\section{MobileCopilot}
\label{sec:mobilecopilot}

MobileCopilot couples an LLM coding agent with the MobileDSL compiler for iterative robot development (Fig.~\ref{fig:pipeline}). The agent selects components and specifies their assembly and behavior in a MobileDSL program; the compiler derives assembly transforms and dependent configuration, checks constraints across modules, and returns diagnostics for program revision.

\subsection{Component Selection and Program Generation}
The agent searches the component library using text, images, or both. Text and image queries are encoded with SigLIP~2~\cite{tschannen2025siglip2}. Each component is represented by front, side, top, and isometric images. Its score is the highest cosine similarity across these views; when both query types are provided, their scores are averaged. Matches to component names and descriptions are also returned. The implementation uses the ViT-B/16 encoder at 384-pixel input resolution.

Search results include descriptions and images for selecting parts, together with geometry and connector frames for choosing mounting positions. With language documentation and an example project, the agent writes the four language modules and links them through shared names.

\subsection{Compiler-Guided Refinement}
The compiler checks the program against the assembly and behavior rules in Sec.~\ref{sec:mobiledsl}. Diagnostics identify a violated constraint and, where available, the source location or a suggested correction. The agent edits the program and recompiles. In the benchmark, agents are instructed to finish once compilation or the baseline's static checks pass. Passing a check does not automatically terminate the run; the agent may continue revising until it decides to finish or reaches the 1,800-s limit. No fixed revision count is imposed.

Figure~\ref{fig:pipeline} illustrates a cross-DSL repair: a task refers to \texttt{@world/wp2}, but the world does not define that goal. The agent adds the missing WorldDSL declaration and compiles again. Other diagnostics can require changing a wheel placement or adding a sensor needed by the chosen navigation setup.

Successful compilation produces the files needed to execute the design in the selected target environment: robot and world models, controller and navigation settings, a behavior tree, and an evaluation plan. The compiler calculates assembly transforms and dependent settings, while the agent chooses the parts and behavior.

\subsection{Simulation and Feedback}
During interactive development, simulator observations and the execution report can guide revisions by a user or an agent. In this benchmark, agents use compiler feedback, including static checks of EvaluateDSL plans, but receive no simulation feedback during authoring. After authoring, submitted programs are scored by the shared runtime evaluator without executing the authored evaluation plans. This protocol measures compiler-guided construction before simulation-based repair.

\section{Evaluation Protocol}
\label{sec:benchmark}
We measure whether generated robot systems execute their programmed behavior in their generated worlds, together with the agent effort required to produce them. Execution is assessed against the generated program rather than an independent specification of the original request.

\subsection{Design Tasks and Authoring}
The benchmark contains 50 tasks: 25 text-only requests and 25 with robot images. Thirty-two tasks specify basic motion, including forward and backward movement, turns, waits, and repetition; 18 include obstacle avoidance or waypoint navigation. The reference designs span two- and four-wheel differential drive, four-wheel mecanum, and three- and four-wheel omni layouts. Each request describes the robot, environment, behavior, and evaluation requirements.

\Needspace{5\baselineskip}
Each task fixes a common execution setting: odometry for elementary motion or laser-based localization in a static map for navigation. The agent chooses the robot, sensors, world, and behavior. MobileDSL derives the map from WorldDSL; the baseline authors it directly.

The agent experiments use Harbor~\cite{harbor} for isolated authoring environments with format-specific instructions, examples, and read-only access to the component library. Agents inspect parts and use static tools without simulation feedback. Submitted programs are compiled or checked before running in the same simulation environment.

\subsection{Execution Checks}
Separate simulation runs test basic driving and completion of the programmed behavior.

\textbf{Motion response (\%)} tests whether velocity commands can drive the assembled robot. We apply a forward velocity of $0.2\,\mathrm{m/s}$ and an angular velocity of $0.4\,\mathrm{rad/s}$ for five simulated seconds. A robot passes if it receives the command and moves at least $0.1\,\mathrm{m}$ or rotates by at least $0.3\,\mathrm{rad}$, with control and simulation operating correctly. Motion is measured from simulator pose.

\textbf{System execution success (\%)} tests whether the robot completes the behavior defined in its generated program. Incorrect controller settings, for example, can prevent path following even when the behavior tree is valid. The behavior tree and applicable navigation actions must succeed, with observations confirming each action.

Navigation actions must end within $0.15\,\mathrm{m}$ of the final goal and, for multi-waypoint actions, pass within $0.3\,\mathrm{m}$ of each waypoint in order. Locomotion-only behaviors are judged through their motion steps. All behaviors require collision avoidance, completion within the task budget, healthy control and simulation, and the required observations.

The common evaluator uses simulator-state traces, action results, and execution events, separately from the generated EvaluateDSL plan. Rates use all \NumCases\ attempts per condition; generation failures and missing execution evidence do not count as successes.

\subsection{Authoring Effort}
We measure the code, model interaction, time, and cost of the full authoring attempt, including revisions. Subsequent simulation and visual evaluation are excluded.

\textbf{Authored LOC} counts nonblank model-written source lines, excluding compiler output and existing library code. \textbf{Generation time} includes model calls, component retrieval, and static validation. \textbf{API cost} is calculated from recorded tokens at the provider rates recorded on 2026-09-11, including cache pricing.

\textbf{Input tokens at last call} records the final call's input length. \textbf{Tokens across all calls} accumulates input and output throughout authoring, including repeated context. \textbf{Total tokens} sums input and output. Cached input is included in total input; uncached input is the remainder. Counts are in thousands (k).

Effort values are per-attempt means. Differences between success rates are reported in percentage points; table arrows indicate the preferred direction.

\section{Experiments}
\label{sec:experiments}
We evaluate whether MobileDSL and its compiler improve system execution success while reducing agent authoring cost. The experiments cover complete-system development, assembly and configuration with fixed components, and ablations that replace individual DSL modules with directly authored files.

\begin{table*}[t]
\centering
\CompactTableCaption
\begin{threeparttable}
\setlength{\tabcolsep}{1.4pt}
\renewcommand{\arraystretch}{1.16}
\caption{\textbf{Simulation success and generation effort with and without MobileDSL.} Bold marks the better value within each agent/model pair.}
\label{tab:main-results}
\begingroup
\fontsize{7.6}{9.0}\selectfont
\begin{tabular*}{\textwidth}{@{\extracolsep{\fill}}cllrrrrrrrrrr@{}}
\toprule
\multirow{2}{*}[-8pt]{\textbf{Agent}} & \multirow{2}{*}[-8pt]{\textbf{Model}} & \multirow{2}{*}[-8pt]{\textbf{Representation}} & \multicolumn{2}{c}{\textbf{Simulation success}} & \multirow{2}{*}[-8pt]{\shortstack{\textbf{Authored}\\\textbf{LOC} $\downarrow$}} & \multirow{2.5}{*}[-5.4pt]{\shortstack{\textbf{Input tokens}\\\textbf{at last call}\\{\scriptsize[k]}}} & \multicolumn{4}{c}{\textbf{Tokens across all calls}} & \multicolumn{2}{c}{\textbf{Generation}}\\
\cmidrule(lr){4-5}\cmidrule(lr){8-11}\cmidrule(l){12-13}
& & & \multicolumn{1}{c}{\shortstack{Motion\\response\\{\scriptsize[\%] $\uparrow$}}} & \multicolumn{1}{c}{\shortstack{System\\execution\\{\scriptsize[\%] $\uparrow$}}} & & & \multicolumn{1}{c}{\shortstack{Total\\input\\{\scriptsize[k] $\downarrow$}}} & \multicolumn{1}{c}{\shortstack{Cached\\input\\{\scriptsize[k]}}} & \multicolumn{1}{c}{\shortstack{Uncached\\input\\{\scriptsize[k]}}} & \multicolumn{1}{c}{\shortstack{Output\\{\scriptsize[k] $\downarrow$}}} & \multicolumn{1}{c}{\shortstack{Time\\{\scriptsize[s] $\downarrow$}}} & \multicolumn{1}{c}{\shortstack{Cost\\{\scriptsize[\$] $\downarrow$}}}\\
\midrule
\multirow{4}{*}[-2pt]{Codex} & \multirow{2}{*}{gpt-5.6-sol} & MobileDSL & \textbf{98} & \textbf{74} & \textbf{43.7} & 47.16 & \textbf{659.43} & 611.40 & 48.03 & \textbf{5.65} & \textbf{285} & \textbf{0.550}\\
& & ROS/Gazebo & 56 & 24 & 505.3 & 79.63 & 1,793.43 & 1,704.40 & 89.03 & 18.37 & 731 & 1.406\\
\cmidrule(lr){2-13}
& \multirow{2}{*}{gpt-6-astra} & MobileDSL & \textbf{100} & \textbf{82} & \textbf{46.8} & 39.06 & \textbf{419.05} & 382.84 & 36.21 & \textbf{2.52} & \textbf{151} & \textbf{0.871}\\
& & ROS/Gazebo & 78 & 40 & 1,301.9 & 50.33 & 569.17 & 520.24 & 48.93 & 5.62 & 250 & 1.291\\
\midrule
\multirow{2}{*}{\shortstack{Claude\\Code}} & \multirow{2}{*}{\shortstack[l]{claude-opus-\\4-8}} & MobileDSL & \textbf{96} & \textbf{64} & \textbf{41.5} & 64.93 & \textbf{1,202.82} & 1,149.62 & 53.20 & \textbf{15.14} & \textbf{268} & \textbf{1.485}\\
& & ROS/Gazebo & 58 & 22 & 870.0 & 100.25 & 2,227.61 & 2,138.53 & 89.08 & 39.95 & 569 & 2.959\\
\bottomrule
\end{tabular*}
\endgroup
\begin{tablenotes}[flushleft]
\fontsize{7.2}{8.6}\selectfont
\item[] \textit{Note.} Cached input is included in total input. Cost is defined (Sec.~\ref{sec:benchmark}).
\end{tablenotes}
\end{threeparttable}
\end{table*}

\begin{figure}[!t]
\centering
\setlength{\abovecaptionskip}{2pt}
\includegraphics[width=\linewidth]{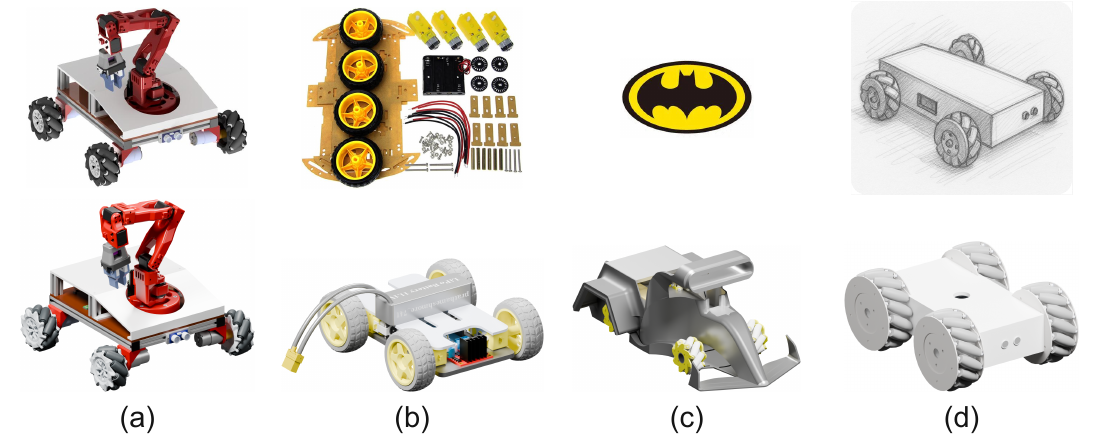}
\caption{\textbf{Robot designs generated with MobileDSL.} Input images (top) and corresponding generated robot designs (bottom).}
\label{fig:main_qualitative}
\end{figure}

\begin{figure*}[!t]
\centering
\setlength{\abovecaptionskip}{2pt}
\includegraphics[width=\linewidth]{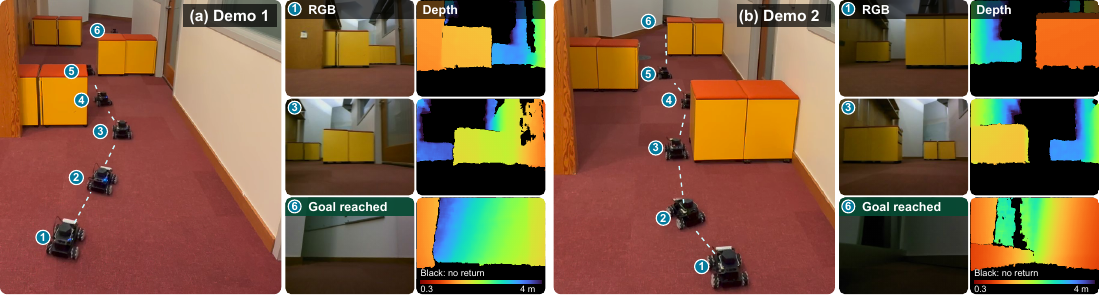}
\caption{\textbf{Real-world deployment with MobileDSL.} Physical robots executing navigation programs generated from MobileDSL while avoiding obstacles.}
\label{fig:real_navigation}
\end{figure*}

\subsection{Experimental Setup}
\label{ex:mobilecopilot}
\textbf{Agent-based development.}
We compare \AgentName\ with \AgentModel\ or gpt-6-astra, and Claude Code with claude-opus-4-8. For each configuration, the agent processes all \NumCases\ requests once using MobileDSL and once by directly authoring ROS/Gazebo files. A run may contain multiple model calls and revisions, all within the same \AgentBudget\,s budget.

Both conditions receive the same requests, images, and component library, with documentation and examples for their formats. Both can search for parts, inspect geometry and connectors, view static robot renderings, and revise their designs. Baseline checks identify malformed or missing files, invalid behavior trees, and unsupported controllers or inconsistent task settings. MobileDSL additionally assembles the robot and generates dependent settings. Both conditions use the same execution environment and collision instrumentation, without simulation feedback during authoring.

\textbf{Fixed-component descriptions.}
\label{sec:coredsl_ex}
We compare CoreDSL, URDF, and Xacro with gpt-5.6-sol and claude-opus-4-8, fixing the selected parts to separate assembly and configuration from retrieval. Every condition receives the same component inventory, dimensions, connectors, and joint records, together with a morphology request or image, documentation, and an unrelated example. Reference assembly programs and placements are withheld. Each model produces one response per case and language without tools or revision, measuring initial authoring rather than feedback-based repair. CoreDSL delegates assembly and configuration to the compiler; URDF and Xacro require explicit robot and controller descriptions, with macros permitted in Xacro. The environment and motion command are supplied.

\textbf{Assembly validity (\%)} measures whether the selected parts form a robot satisfying the supplied mechanical requirements. A checker examines the materialized URDF against shared component records: identities and counts, geometry, connector alignment, and joint types and axes. For example, a wheel must mate with an available chassis mount and use the specified axle direction. All assemblies satisfying these rules count as valid; structure and controller-configuration verdicts are recorded separately. We report this measure alongside motion response and authoring effort, using all \NumCases\ attempts per condition.

\textbf{Individual-DSL ablations.}
\label{sec:ablation}
We compare full MobileDSL, direct file authoring, and four leave-one-DSL-out variants. Replacements are URDF and robot settings for CoreDSL, world/map files and coordinates for WorldDSL, behavior-tree and execution files for TaskDSL, and evaluation-plan JSON for EvaluateDSL. Remaining DSLs retain compiler support and resolve references against these files. Neither the compiler-generated evaluation plans nor their directly authored JSON replacements are executed in this benchmark; all configurations use the same runtime evaluator. Each configuration uses \AblationAgent{} / \AblationModel{} on all \NumCases\ requests once, under the same authoring budget and feedback restrictions as the agent-based comparison.

The execution and effort measures follow Sec.~\ref{sec:benchmark}. Unsuccessful runs remain in the success-rate denominators; all model calls and revisions contribute to authoring effort. For fixed-component descriptions, authored LOC includes the robot description and required controller and bridge settings.

We compare each authoring interface together with its assembly and configuration support.

\begin{table*}[t]
\centering
\CompactTableCaption
\setlength{\tabcolsep}{3pt}
\renewcommand{\arraystretch}{1.12}
\caption{\textbf{Robot-description generation with fixed components.} Bold marks the best value within each model.}
\label{tab:rddl-comparison}
\begingroup
\fontsize{7.6}{9.0}\selectfont
\newcommand{\TableHeader}[1]{%
  {\renewcommand{\arraystretch}{1.0}%
  \begin{tabular}[c]{@{}c@{}}#1\end{tabular}}}
\begin{tabular*}{\textwidth}{@{\extracolsep{\fill}}*{9}{c}@{}}
\toprule
\textbf{Model} & \textbf{Language} &
\TableHeader{\textbf{Motion response}\\{[\%] $\uparrow$}} &
\TableHeader{\textbf{Assembly validity}\\{[\%] $\uparrow$}} &
\TableHeader{\textbf{Authored}\\\textbf{LOC} $\downarrow$} &
\TableHeader{\textbf{Input tokens}\\{[k] $\downarrow$}} &
\TableHeader{\textbf{Output tokens}\\{[k] $\downarrow$}} &
\TableHeader{\textbf{Time}\\{[s] $\downarrow$}} &
\TableHeader{\textbf{Cost}\\{[\$] $\downarrow$}}\\
\midrule
\multirow{3}{*}{gpt-5.6-sol} & URDF & 92 & 96 & 313.3 & 10.53 & 5.42 & 68.5 & 0.1611\\
& Xacro & 94 & 98 & 302.7 & 10.57 & 5.12 & 67.9 & 0.1552\\
& CoreDSL & \textbf{96} & \textbf{100} & \textbf{10.7} & \textbf{6.58} & \textbf{0.53} & \textbf{9.5} & \textbf{0.0434}\\
\midrule
\multirow{3}{*}{claude-opus-4-8} & URDF & 92 & 88 & 389.0 & 14.85 & 11.38 & 112.8 & 0.3588\\
& Xacro & \textbf{94} & 86 & 393.7 & 14.95 & 11.44 & 113.5 & 0.3606\\
& CoreDSL & 88 & \textbf{92} & \textbf{10.7} & \textbf{9.04} & \textbf{1.20} & \textbf{13.7} & \textbf{0.0752}\\
\bottomrule
\end{tabular*}
\endgroup
\par\vspace{12pt}
\end{table*}

\begin{table}[t]
\centering
\CompactTableCaption
\caption{\textbf{Effect of individual DSLs on simulation success and generation effort.} \textbf{Bold} and \underline{underline} mark the best and second-best values, respectively, including ties.}
\label{tab:case-metrics}
\begingroup
\fontsize{7.8}{9.4}\selectfont
\setlength{\tabcolsep}{2.2pt}
\renewcommand{\arraystretch}{1.15}
\newcommand{\TableHeader}[1]{%
  {\renewcommand{\arraystretch}{1.0}%
  \begin{tabular}[c]{@{}c@{}}#1\end{tabular}}}
\begin{tabular*}{\columnwidth}{@{\extracolsep{\fill}}*{7}{c}@{}}
\toprule
\textbf{Configuration} &
\TableHeader{\textbf{Motion}\\\textbf{response}\\{[\%] $\uparrow$}} &
\TableHeader{\textbf{System}\\\textbf{execution}\\{[\%] $\uparrow$}} &
\TableHeader{\textbf{Authored}\\\textbf{LOC} $\downarrow$} &
\TableHeader{\textbf{Total}\\\textbf{tokens}\\{[k] $\downarrow$}} &
\TableHeader{\textbf{Time}\\{[s] $\downarrow$}} &
\TableHeader{\textbf{Cost}\\{[\$] $\downarrow$}}\\
\midrule
\textbf{Full MobileDSL} & \textbf{98} & \textbf{74} & \textbf{43.7} & \underline{665} & 285 & \textbf{0.55}\\
No CoreDSL         & 84 & 36 & 333.4 & 1,126 & 368 & 0.96\\
No WorldDSL        & \textbf{98} & \underline{70} & 135.1 & 742 & \underline{246} & 0.64\\
No TaskDSL         & 92 & \textbf{74} & 170.2 & 1,129 & 272 & 0.89\\
No EvaluateDSL     & \underline{96} & \underline{70} & \underline{89.7} & \textbf{650} & \textbf{206} & \underline{0.56}\\
ROS/Gazebo files   & 56 & 24 & 505.3 & 1,812 & 731 & 1.41\\
\bottomrule
\end{tabular*}
\endgroup
\par\vspace{8pt}
\end{table}

\MCResultsStart
\subsection{Results and Q\&As}

\MCQuestion{1}{Do more generated systems complete their programmed behavior with MobileDSL?}
\MCAnswer{Yes---MobileDSL helps agents produce systems that carry out the behavior defined in their programs.}

\begingroup
\setlength{\intextsep}{4pt}
\setlength{\columnsep}{8pt}
\begin{wrapfigure}[10]{r}{0.42\columnwidth}
\centering
\setlength{\abovecaptionskip}{2pt}
\includegraphics[width=\linewidth]{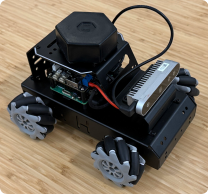}
\caption{Physical robot platform.}
\label{fig:robot_closeup}
\end{wrapfigure}
Table~\ref{tab:main-results} shows higher motion response and system execution success across the tested agent configurations. Execution success increases by 42--50 percentage points, so the benefit extends beyond producing a robot that moves: more designs connect the robot model, controller, navigation, and behavior well enough to complete the actions defined in their generated programs. The gap between motion response and execution success shows why basic mobility alone is insufficient to assess integration. Figure~\ref{fig:main_qualitative}(a)--(d) illustrates the generated robot designs. Figure~\ref{fig:real_navigation} shows programs generated from MobileDSL running on physical robots. Figure~\ref{fig:robot_closeup} shows the physical platform. Please refer to our video for the physical robot demonstrations.
\par\endgroup

\MCQuestion{2}{Does this improvement require more authoring effort?}
\MCAnswer{No---MobileDSL reduces the work needed to develop a robot system rather than requiring extra effort to obtain better results.}

Authored code decreases by approximately 12--28 times, together with generation time and API cost (Table~\ref{tab:main-results}). Accumulated input and output tokens also decrease, showing that the savings extend beyond shorter final files to less model interaction during construction and repair.

\MCQuestion{3}{Can CoreDSL simplify robot assembly and configuration?}
\MCAnswer{Yes---CoreDSL reduces the work of assembling a robot and keeping its control settings consistent with its structure.}

With the same components supplied to every language and no revision, CoreDSL achieves the highest assembly validity with both models while requiring approximately 28--37 times less code and 7--8 times less generation time (Table~\ref{tab:rddl-comparison}). These results locate a substantial benefit in expressing the assembly and its dependent settings, rather than in retrieval or iterative repair. Authors can leave component alignment and dependent configuration to the compiler instead of reconstructing them in separate descriptions.

In the single-response setting, claude-opus-4-8 violated CoreDSL's documented wheel-naming conventions in four cases, whereas gpt-5.6-sol followed them. This suggests differences in how reliably models apply newly introduced DSL conventions without compiler-guided revision.

\MCQuestion{4}{Which DSL contributes most to system execution success?}
\MCAnswer{CoreDSL contributes most in the tested workflow, through its support for assembling the robot and configuring it to operate.}

Replacing CoreDSL reduces execution success from 74\% to 36\% and increases authored code more than sevenfold (Table~\ref{tab:case-metrics}). This is the largest change among the individual-DSL replacements. Together with the fixed-component comparison, it points to assembly and robot-dependent configuration as a major source of the observed benefit. The remaining DSLs cannot by themselves remove the work of keeping an explicitly authored robot model and its settings consistent.

\MCQuestion{5}{What do WorldDSL, TaskDSL, and EvaluateDSL add?}
\MCAnswer{They simplify authoring of the environment, behavior, and requirements, with different tradeoffs in the work needed to generate a program.}

WorldDSL and TaskDSL show their clearest benefits in authoring (Table~\ref{tab:case-metrics}). Replacing WorldDSL yields a similar success rate with more source code; replacing TaskDSL retains the full system's success rate but requires more code, tokens, and cost. They therefore provide more compact ways to describe the environment and behavior, even when execution success changes little.

EvaluateDSL shows a different tradeoff: replacing it increases authored code but decreases total tokens and generation time. A shorter final description need not be faster to produce. Because the runtime evaluator is unchanged, this result concerns requirement authoring rather than removal of runtime checks. The 74\% versus 70\% success difference is only two cases, so it provides no firm reliability conclusion on its own.

\section{Limitations}
\label{sec:limitations}
\paragraph{Component Library}
MobileDSL assembles robots from reusable component descriptions. Its component coverage is therefore bounded by the annotated library. Extending this library requires preparing and validating geometry, connector frames, and physical parameters for additional parts. These records are then reused across designs, making their preparation part of library expansion rather than each new assembly.

\paragraph{Development Workflows}
The benchmark evaluates compiler-guided authoring under common navigation settings, before simulation-based revision. Additional comparisons can assess how runtime feedback changes the work required over successive design iterations. The hardware examples in Fig.~\ref{fig:real_navigation} complement the simulation comparisons with qualitative demonstrations; repeated trials would quantify variation across platforms and operating conditions.

\clearpage
\section{Discussion and Conclusion}

MobileCopilot shows how a coding agent can use a domain-specific language and compiler as tools for mobile robot development. Through MobileDSL, the agent selects components and specifies their assembly and behavior, while the compiler derives dependent settings and checks cross-module constraints across robot structure, environment, behavior, and evaluation requirements. On our 50-task benchmark, this workflow increases system execution success by 42--50 percentage points over direct ROS/Gazebo authoring while reducing API cost by approximately 33--61\%. These results suggest that a coding agent's effectiveness depends not only on its model, but also on the languages and tools it uses. More broadly, better coding agents can come from better languages, not just better context.

\makeatletter
\if@firstcolumn
  \balance
\else
  \AddToHookNext{shipout/after}{\balance}
\fi
\makeatother


\begin{thebibliography}{10}
\providecommand{\url}[1]{#1}
\csname url@rmstyle\endcsname
\providecommand{\newblock}{\relax}
\providecommand{\bibinfo}[2]{#2}
\providecommand\BIBentrySTDinterwordspacing{\spaceskip=0pt\relax}
\providecommand\BIBentryALTinterwordstretchfactor{4}
\providecommand\BIBentryALTinterwordspacing{\spaceskip=\fontdimen2\font plus
\BIBentryALTinterwordstretchfactor\fontdimen3\font minus
  \fontdimen4\font\relax}
\providecommand\BIBforeignlanguage[2]{{%
\expandafter\ifx\csname l@#1\endcsname\relax
\typeout{** WARNING: IEEEtran.bst: No hyphenation pattern has been}%
\typeout{** loaded for the language `#1'. Using the pattern for}%
\typeout{** the default language instead.}%
\else
\language=\csname l@#1\endcsname
\fi
#2}}

\bibitem{ramdani2019safe}
N.~Ramdani, A.~Panayides, M.~Karamousadakis, M.~Mellado, R.~Lopez,
  C.~Christophorou, M.~Rebiai, M.~Blouin, E.~Vellidou, and D.~Koutsouris, ``A
  safe, efficient and integrated indoor robotic fleet for logistic applications
  in healthcare and commercial spaces: the endorse concept,'' in \emph{2019
  20th IEEE international conference on mobile data management (MDM)}.\hskip
  1em plus 0.5em minus 0.4em\relax IEEE, 2019, pp. 425--430.

\bibitem{bloss2011mobile}
R.~Bloss, ``Mobile hospital robots cure numerous logistic needs,''
  \emph{Industrial Robot: An International Journal}, vol.~38, no.~6, pp.
  567--571, 2011.

\bibitem{kebede2024review}
G.~A. Kebede, A.~A. Gelaw, H.~Andualem, and A.~T. Hailu, ``Review of the
  characteristics of mobile robots for health care application,''
  \emph{International Journal of Intelligent Robotics and Applications},
  vol.~8, no.~2, pp. 480--502, 2024.

\bibitem{garcia2020robotics}
S.~Garc{\'\i}a, D.~Str{\"u}ber, D.~Brugali, T.~Berger, and P.~Pelliccione,
  ``Robotics software engineering: A perspective from the service robotics
  domain,'' in \emph{Proceedings of the 28th ACM Joint Meeting on European
  Software Engineering Conference and Symposium on the Foundations of Software
  Engineering}, 2020, pp. 593--604.

\bibitem{garaffa2021reinforcement}
L.~C. Garaffa, M.~Basso, A.~A. Konzen, and E.~P. de~Freitas, ``Reinforcement
  learning for mobile robotics exploration: A survey,'' \emph{IEEE Transactions
  on Neural Networks and Learning Systems}, vol.~34, no.~8, pp. 3796--3810,
  Aug. 2023.

\bibitem{raj2022comprehensive}
R.~Raj and A.~Kos, ``A comprehensive study of mobile robot: history,
  developments, applications, and future research perspectives,'' \emph{Applied
  Sciences}, vol.~12, no.~14, p. 6951, 2022.

\bibitem{urdf}
\BIBentryALTinterwordspacing
{ROS.org}, ``Unified robot description format (urdf),'' 2010. [Online].
  Available: \url{http://wiki.ros.org/urdf}
\BIBentrySTDinterwordspacing

\bibitem{todorov2012mujoco}
E.~Todorov, T.~Erez, and Y.~Tassa, ``Mujoco: A physics engine for model-based
  control,'' in \emph{2012 IEEE/RSJ international conference on intelligent
  robots and systems}.\hskip 1em plus 0.5em minus 0.4em\relax IEEE, 2012, pp.
  5026--5033.

\bibitem{ros2009}
\BIBentryALTinterwordspacing
M.~Quigley, B.~Gerkey, K.~Conley, J.~Faust, T.~Foote, J.~Leibs, E.~Berger,
  R.~Wheeler, and A.~Ng, ``{ROS}: an open-source robot operating system,'' in
  \emph{Proceedings of the IEEE International Conference on Robotics and
  Automation (ICRA) Workshop on Open Source Robotics}, 2009. [Online].
  Available: \url{https://ai.stanford.edu/~ang/papers/icraoss09-ROS.pdf}
\BIBentrySTDinterwordspacing

\bibitem{gazebo2025}
{Open Robotics}, ``Gazebo - robot simulation platform, home page,''
  \url{https://gazebosim.org/home}, 2025, online; accessed 15 September 2025.

\bibitem{ogren2022behavior}
P.~{\"O}gren and C.~I. Sprague, ``Behavior trees in robot control systems,''
  \emph{Annual Review of Control, Robotics, and Autonomous Systems}, vol.~5,
  no.~1, pp. 81--107, 2022.

\bibitem{guiochet2017safety}
J.~Guiochet, M.~Machin, and H.~Waeselynck, ``Safety-critical advanced robots: A
  survey,'' \emph{Robotics and Autonomous Systems}, vol.~94, pp. 43--52, 2017.

\bibitem{tola2024understanding}
D.~Tola and P.~Corke, ``Understanding urdf: A dataset and analysis,''
  \emph{IEEE Robotics and Automation Letters (RA-L)}, vol.~9, no.~5, pp.
  4479--4486, 2024.

\bibitem{zeng2022codeaspolicies}
J.~Liang, W.~Huang, F.~Xia, P.~Xu, K.~Hausman, B.~Ichter, P.~Florence, and
  A.~Zeng, ``Code as policies: Language model programs for embodied control,''
  \emph{arXiv preprint arXiv:2209.07753}, 2022.

\bibitem{hu2024codebotler}
Z.~Hu, F.~Lucchetti, C.~Schlesinger, Y.~Saxena, A.~Freeman, S.~Modak, A.~Guha,
  and J.~Biswas, ``Deploying and evaluating llms to program service mobile
  robots,'' \emph{IEEE Robotics and Automation Letters}, vol.~9, no.~3, pp.
  2853--2860, 2024.

\bibitem{aizawa2025tools}
\BIBentryALTinterwordspacing
K.~Aizawa, ``Writing effective tools for agents---with agents,'' Anthropic
  Engineering, Sept. 2025, september 11. [Online]. Available:
  \url{https://www.anthropic.com/engineering/writing-tools-for-agents}
\BIBentrySTDinterwordspacing

\bibitem{mernik2005and}
M.~Mernik, J.~Heering, and A.~M. Sloane, ``When and how to develop
  domain-specific languages,'' \emph{ACM computing surveys (CSUR)}, vol.~37,
  no.~4, pp. 316--344, 2005.

\bibitem{nordmann2014survey}
A.~Nordmann, N.~Hochgeschwender, and S.~Wrede, ``A survey on domain-specific
  languages in robotics,'' in \emph{International conference on simulation,
  modeling, and programming for autonomous robots}.\hskip 1em plus 0.5em minus
  0.4em\relax Springer, 2014, pp. 195--206.

\bibitem{xacro}
\BIBentryALTinterwordspacing
{ROS.org}, ``Xacro: Xml macros,'' 2010. [Online]. Available:
  \url{http://wiki.ros.org/xacro}
\BIBentrySTDinterwordspacing

\bibitem{koenig2004gazebo}
N.~Koenig and A.~Howard, ``Design and use paradigms for gazebo, an open-source
  multi-robot simulator,'' \emph{Proceedings of the IEEE/RSJ International
  Conference on Intelligent Robots and Systems (IROS)}, pp. 2149--2154, 2004.

\bibitem{nvidia2023openusd}
\BIBentryALTinterwordspacing
A.~Luk, P.~Lee, and R.~Gasoto, ``Using {OpenUSD} for modular and scalable
  robotic simulation and development,'' NVIDIA Technical Blog, Mar. 2025.
  [Online]. Available:
  \url{https://developer.nvidia.com/blog/using-openusd-for-modular-and-scalable-robotic-simulation-and-development/}
\BIBentrySTDinterwordspacing

\bibitem{colledanchise2018behavior}
M.~Colledanchise and P.~{\"O}gren, \emph{Behavior trees in robotics and AI: An
  introduction}.\hskip 1em plus 0.5em minus 0.4em\relax CRC Press, 2018.

\bibitem{iovino2022survey}
M.~Iovino, E.~Scukins, J.~Styrud, P.~{\"O}gren, and C.~Smith, ``A survey of
  behavior trees in robotics and ai,'' \emph{Robotics and Autonomous Systems},
  vol. 154, p. 104096, 2022.

\bibitem{dhouib2012robotml}
S.~Dhouib, S.~Kchir, S.~Stinckwich, T.~Ziadi, and M.~Ziane, ``Robotml, a
  domain-specific language to design, simulate and deploy robotic
  applications,'' in \emph{Proceedings of the International Conference on
  Simulation, Modeling, and Programming for Autonomous Robots (SIMPAR)}.\hskip
  1em plus 0.5em minus 0.4em\relax Springer, 2012, pp. 149--160.

\bibitem{isaac_sim2025}
{NVIDIA Corporation}, ``Isaac sim (version 5.0) — robotics simulation
  platform on nvidia omniverse,'' \url{https://github.com/isaac-sim/IsaacSim},
  2025, accessed: 15 September 2025.

\bibitem{zhao2020robogrammar}
A.~Zhao, J.~Xu, M.~Konakovi{\'c}~Lukovi{\'c}, J.~Hughes, A.~Spielberg, D.~Rus,
  and W.~Matusik, ``{RoboGrammar}: Graph grammar for terrain-optimized robot
  design,'' \emph{ACM Transactions on Graphics}, vol.~39, no.~6, pp. 1--16,
  2020.

\bibitem{ringel2025text2robot}
R.~P. Ringel, Z.~S. Charlick, J.~Liu, B.~Xia, and B.~Chen, ``Text2robot:
  Evolutionary robot design from text descriptions,'' in \emph{2025 IEEE
  International Conference on Robotics and Automation (ICRA)}.\hskip 1em plus
  0.5em minus 0.4em\relax IEEE, 2025, pp. 5789--5797.

\bibitem{khan2025evolution}
M.~H. Khan, A.~Myshlyaev, A.~Lykov, M.~A. Cabrera, and D.~Tsetserukou,
  ``Evolution 6.0: Evolving robotic capabilities through generative design,''
  \emph{arXiv preprint arXiv:2502.17034}, 2025.

\bibitem{song2025laser}
J.~Song, Y.~Yang, H.~Xiao, W.~Peng, W.~Yao, and F.~Wang, ``Laser: Towards
  diversified and generalizable robot design with large language models,'' in
  \emph{The Thirteenth International Conference on Learning Representations},
  2025.

\bibitem{fang2025robomore}
J.~Fang, Y.~Sun, C.~Ma, Q.~Lu, and L.~Yao, ``Robomore: Llm-based robot
  co-design via joint optimization of morphology and reward,'' \emph{arXiv
  preprint arXiv:2506.00276}, 2025.

\bibitem{gazebo_models}
{Open Source Robotics Foundation}, ``Gazebo models,''
  \url{https://github.com/osrf/gazebo_models}.

\bibitem{gazebo_fuel}
------, ``Gazebo fuel models,'' \url{https://app.gazebosim.org/fuel/models}.

\bibitem{macenski2020marathon2}
\BIBentryALTinterwordspacing
S.~Macenski, F.~Martín, R.~White, and J.~Ginés~Clavero, ``The marathon 2: A
  navigation system,'' in \emph{2020 IEEE/RSJ International Conference on
  Intelligent Robots and Systems (IROS)}, 2020. [Online]. Available:
  \url{https://github.com/ros-planning/navigation2}
\BIBentrySTDinterwordspacing

\bibitem{tschannen2025siglip2}
M.~Tschannen, A.~Gritsenko, X.~Wang, M.~F. Naeem, I.~Alabdulmohsin,
  N.~Parthasarathy, T.~Evans, L.~Beyer, Y.~Xia, B.~Mustafa, O.~H{\'e}naff,
  J.~Harmsen, A.~Steiner, and X.~Zhai, ``{SigLIP} 2: Multilingual
  vision-language encoders with improved semantic understanding, localization,
  and dense features,'' \emph{arXiv preprint arXiv:2502.14786}, 2025.

\bibitem{harbor}
{Harbor}, ``{Harbor}: Core concepts,''
  \url{https://www.harborframework.com/docs/core-concepts}, accessed: September
  13, 2026.

\end{thebibliography}
\end{document}